\documentclass[11pt,a4paper]{article}
\usepackage[table]{xcolor}
\usepackage{times,latexsym}

\usepackage{url}
\usepackage[T1]{fontenc}
\usepackage[acceptedWithA]{tacl2018v2}

\usepackage{verbatim}
\usepackage{smartdiagram}
\usepackage{etoolbox}
\usepackage{tikz}
\usepackage{amsmath}    
\usepackage{multirow}   
\usepackage{algorithm}    
\usepackage{amsfonts}   
\usepackage{lipsum}   
\usepackage{framed}   
\usepackage{graphicx}   
\usepackage{longtable}    
\usepackage{caption}    
\usepackage{subcaption}   
\usepackage{graphics}
\usepackage{float}
\usepackage{stfloats}
\usepackage{bm}
\usepackage{amssymb}
\usepackage{pifont}
\usepackage{mwe}
\usepackage[percent]{overpic}
\usepackage[normalem]{ulem}

\newif\ifcomment
\commenttrue

\usepackage{framed}
\usepackage{mdwlist}
\usepackage{latexsym}
\usepackage{xcolor}
\usepackage{nicefrac}
\usepackage{booktabs}
\usepackage{amsfonts}
\usepackage{bold-extra}
\usepackage{amsmath}
\usepackage{dsfont}
\usepackage{amssymb}
\usepackage{bm}
\usepackage{graphicx}
\usepackage{mathtools}
\usepackage{microtype}
\usepackage{multirow}
\usepackage{multicol}
\usepackage{latexsym,comment}

\newcommand{\gem}[1]{\mbox{\textsc{gem}}}
\newcommand{\abr}[1]{\textsc{#1}}

\newcommand{\email}[1]{ {\small \href{mailto://#1}{\texttt{#1} }  }}
\newcommand{\emaillink}[2]{ {\small \href{mailto://#2}{\texttt{#1} }  }}

\newcommand{\hidetext}[1]{}
\newcommand{\ignore}[1]{}

\ifcomment
\newcommand{\pinaforecomment}[3]{\colorbox{#1}{\parbox{.8\linewidth}{#2: #3}}}
\else
\newcommand{\pinaforecomment}[3]{}
\fi

\newcommand{\smallurl}[1]{ \begin{tiny}\url{#1}\end{tiny}}

\definecolor{lightblue}{HTML}{3cc7ea}
\definecolor{CUgold}{HTML}{CFB87C}
\definecolor{grey}{rgb}{0.95,0.95,0.95}
\definecolor{ceil}{rgb}{0.57, 0.63, 0.81}

\newcommand{\qb}[0]{Quizbowl}

\newcommand{\mb}[1]{\boldsymbol{\mathbf{#1}}}
\newcommand{\ntotalquestions}[0]{1213}
\newcommand{\ntotalsentences}[0]{6,541}
\newcommand{\ntotalwriters}[0]{115}
\newcommand{\nansweroptions}[0]{25,000}
\newcommand{\ntotalNeural}[0]{406}
\newcommand{\ntotalIR}[0]{807}
\newcommand{\xmark}{\ding{55}}
\newcommand{\Checkmark}{\ding{51}}
\newcommand{\challenge}{adversarially-authored}
\definecolor{ao}{rgb}{0.0, 0.5, 0.0}

\definecolor{color0}{HTML}{FFFFFF}
\definecolor{color1}{HTML}{FEE5D8}
\definecolor{color2}{HTML}{FDCAB5}
\definecolor{color3}{HTML}{FCAB8F}
\definecolor{color4}{HTML}{FC8A6A}
\definecolor{color5}{HTML}{FB694A}
\definecolor{color6}{HTML}{F14432}
\definecolor{color7}{HTML}{D92523}
\definecolor{color8}{HTML}{BC141A}
\definecolor{color9}{HTML}{980C13}
\definecolor{gitred}{HTML}{FDB8C0}
\definecolor{gitgreen}{HTML}{ACF294}
\newcommand*{\mybox}[2]{\tikz[anchor=base,baseline=0pt,rounded corners=0pt, inner sep=0.2mm] \node[fill=#1] (X) {#2};}

\graphicspath{{2018_tacl_trick/images/}}

\definecolor{Goldenrod}{HTML}{3cc7ea}
\definecolor{colorsquad}{rgb}{0,1,0}
\definecolor{colorsnli}{rgb}{1,0,0}
\definecolor{colorvqa}{rgb}{1,1,0}

\title{Trick Me If You Can: Human-in-the-loop Generation of \\ Adversarial Examples for Question Answering}

\author{Eric Wallace \\
       \abr{ee} and \abr{umiacs} \\       
        University of Maryland \\
        \email{ewallac2@umiacs.umd.edu}
        \And
        Pedro Rodriguez, Shi Feng \\
        \abr{cs}, \abr{umiacs} \\
        University of Maryland \\
        \emaillink{\{pedro,}{pedro@umiacs.umd.edu}\emaillink{shifeng\}}{shifeng@umiacs.umd.edu} \\
        \emaillink{@umiacs.umd.edu}{shifeng@umiacs.umd.edu}
                \And
        Ikuya Yamada \\
        Studio Ousia \\
        \email{ikuya@ousia.jp}
        \AND
        Jordan Boyd-Graber \\
        \abr{cs}, iSchool, \abr{umiacs}, \abr{lsc} \\
        University of Maryland \\
        \email{jbg@umiacs.umd.edu}}

\date{}

\begin{document}
\maketitle

\begin{abstract}
Adversarial evaluation stress tests a model's understanding of natural language. While past approaches expose superficial patterns, the resulting adversarial examples are limited in complexity and diversity. We propose human-in-the-loop adversarial generation, where human authors are guided to break models. We aid the authors with interpretations of model predictions through an interactive user interface. We apply this generation framework to a question answering task called Quizbowl, where trivia enthusiasts craft adversarial questions. The resulting questions are validated via live human--computer matches: although the questions appear ordinary to humans, they systematically stump neural and information retrieval models. The adversarial questions cover diverse phenomena from multi-hop reasoning to entity type distractors, exposing open challenges in robust question answering.
\end{abstract}

\section{Introduction}
\label{sec:introduction}

Proponents of machine learning claim human
parity on tasks like reading comprehension~\cite{yu2018qanet}
and commonsense
inference~\cite{devlin2018BERT}. Despite these successes,
many evaluations neglect that computers solve natural language
processing (\abr{nlp}) tasks 
in a fundamentally different way than humans.  

\begin{figure*}[t]
\centering
\includegraphics[width=0.8\linewidth, trim=0.0cm 0.3cm 0cm 0.2cm, clip]{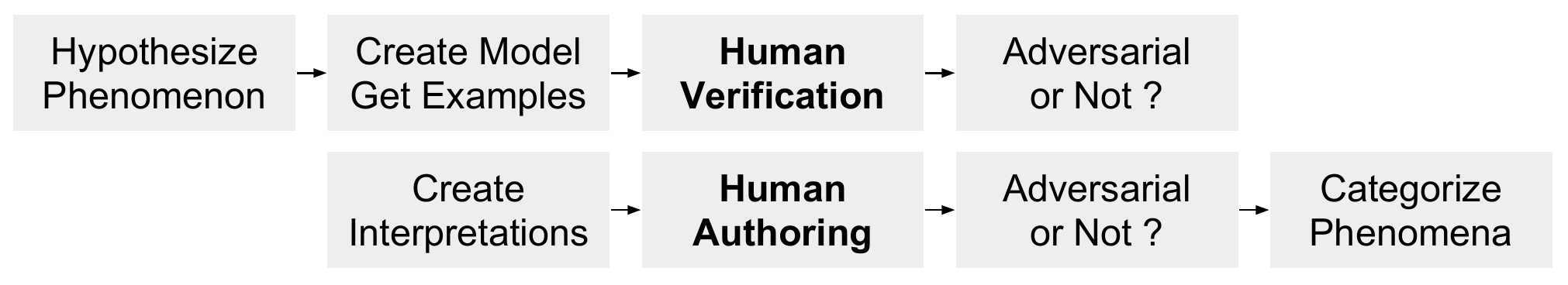}
\caption{Adversarial evaluation in \abr{nlp} typically focuses on a specific
  phenomenon (e.g., word replacements) and then generates the corresponding examples (top).
  Consequently, adversarial examples are limited to
  the diversity of what the underlying generative model or perturbation rule can produce and also require downstream
  human evaluation to ensure validity. Our setup (bottom) instead has
  human-authored examples, using human--computer collaboration to craft adversarial examples with greater diversity.}
\label{fig:flow_chart}
\end{figure*}

Models can succeed without developing ``true'' language understanding,
instead learning superficial patterns from
crawled~\cite{chen2016thorough} or manually annotated
datasets~\cite{kaushik2018reading,gururangan2018annotation}.  Thus,
recent work stress tests models via adversarial evaluation:
elucidating a system's capabilities by exploiting its
weaknesses~\cite{jia2017adversarial,belinkov2019survey}.
Unfortunately, while adversarial evaluation reveals simplistic model
failures~\cite{ribeiro2018sear,mudrakarta2018understand}, exploring
more complex failure patterns requires human involvement (Figure~\ref{fig:flow_chart}):
automatically modifying natural language examples
without invalidating them is difficult. Hence, the diversity
of adversarial examples is often severely restricted.
	
Instead, our human--computer hybrid approach uses human creativity to
generate adversarial examples.  A user interface presents model
interpretations and helps users craft model-breaking examples
(Section~\ref{sec:dataset}).  We apply this to a question answering
(\abr{qa}) task called \qb{}, where trivia enthusiasts---who write
questions for academic competitions---create diverse examples
that stump existing \abr{qa} models.

The \challenge{} test set is nonetheless as easy as
regular questions for humans (Section~\ref{sec:human}), but
the relative accuracy of strong \abr{qa} models drops as much as 40\%
(Section~\ref{sec:experiments}).
We also host live human vs. computer matches---where models typically defeat
top human teams---but observe spectacular model failures
on adversarial questions.

Analyzing the adversarial edits uncovers phenomena
that humans can solve but computers cannot
(Section~\ref{sec:limitations}), validating that our framework
uncovers creative, targeted adversarial edits (Section~\ref{sec:help}).
Our resulting adversarial dataset presents a fun, challenging,
and diverse resource for future \abr{qa} research: a system that masters
it will demonstrate more robust language understanding.
\section{Adversarial Evaluation for \abr{nlp}}
\label{sec:recent}

Adversarial examples~\cite{szegedy2013-intriguing} often reveal
model failures better than traditional test sets. However, automatic
adversarial generation is tricky for \abr{nlp} (e.g., by replacing words) without
changing an example's meaning or invalidating it.

Recent work side-steps this by focusing on simple
transformations that preserve meaning. For instance,
\citet{ribeiro2018sear} generate adversarial
perturbations such as replacing \emph{What has} $\to$
\emph{What's}. Other minor perturbations such as
typos~\cite{belinkov2018synthetic}, adding distractor
sentences~\cite{jia2017adversarial,mudrakarta2018understand}, or
character replacements~\cite{ebrahimi2017hotflip} preserve meaning
while degrading model performance.

Generative models can discover more adversarial perturbations but
require \textit{post hoc} human verification of the examples.
For example, neural paraphrase or language models can
generate syntax modifications~\cite{iyyerscpn2018}, plausible
captions~\cite{zellers2018swag}, or \textsc{nli}
premises~\cite{zhao2017generating}. These methods improve
example-level diversity but mainly target a specific phenomenon, e.g.,
rewriting question syntax.

Furthermore, existing adversarial perturbations are restricted to sentences---not the
paragraph inputs of \qb{} and other
tasks---due to challenges in long-text generation. For instance,
syntax paraphrase networks~\cite{iyyerscpn2018} applied to \qb{} only
yield valid paraphrases 3\% of the time (Appendix~\ref{sec:scpn}).

\subsection{Putting a Human in the Loop}\label{sec:loop}

Instead, we task human authors with \emph{adversarial writing} of questions:
generating examples which break a specific \abr{qa} system but
are still answerable by humans. We expose model predictions and
interpretations to question authors, who find question edits that confuse the model. 

The user interface makes the adversarial writing process interactive and
model-driven, in contrast to adversarial examples written independent of a model~\cite{ettinger2017towards}. The result is an \challenge{} dataset that explicitly exposes a model's limitations by design.

Human-in-the-loop generation can replace or aid model-based
adversarial generation approaches. Creating
interfaces and interpretations is often easier than designing and
training generative models for specific domains. In domains where
adversarial generation is feasible, human creativity
can reveal which tactics automatic approaches can
later emulate. Model-based and human-in-the-loop generation approaches can
also be combined by training models to mimic human adversarial edit
history, using the relative merits of both approaches.
\section{Our QA Testbed: \qb{}}\label{sec:dataset}

The ``gold standard'' of academic competitions between universities
and high schools is \qb{}. Unlike \abr{qa} formats such as
Jeopardy!~\cite{ferruci-10}, \qb{} questions are
designed to be interrupted: questions are read to two competing teams
and whoever knows the answer first interrupts the question and
``buzzes in''.

This style of play requires questions to be structured
``pyramidally''~\cite{jose2017craft}: questions start with difficult clues and get
progressively easier. These
questions are carefully crafted to
allow the most knowledgeable player to answer first. A question on
Paris that begins ``this capital of France'' would test 
reaction speed, not knowledge; thus, skilled authors arrange
the clues so players will recognize them with increasing
probability (Figure~\ref{fig:ex}).

The answers to \qb{} questions are typically well-known entities.  In
the \abr{qa} community~\cite{hirschman-01}, this is called ``factoid''
\abr{qa}: the entities come from a relatively
closed set of possible answers.

\begin{figure}[t!]
\centering
\tikz\node[draw=white!40!black,inner sep=1pt,line width=0.3mm,rounded corners=0.1cm]{
  \begin{tabular}{p{0.46\textwidth}}
    \small
The protagonist of this opera describes the future day when her lover will
arrive on a boat in the aria ``Un Bel Di'' or ``One Beautiful Day''. The only
baritone role in this opera is the consul Sharpless who reads letters for the
protagonist, who has a maid named Suzuki. That protagonist blindfolds her child
Sorrow before stabbing herself when her lover B.\ F.\ Pinkerton returns with a
wife. For 10 points, name this Giacomo Puccini opera about an American
lieutenant's affair with the Japanese woman Cio-Cio San.\\
\textbf{Answer}: \underline{Madama Butterfly}
\end{tabular}
};
\caption{An example \qb{} question. The question becomes progressively
  easier (for humans) to answer later on; thus, more knowledgeable
  players can answer after hearing fewer clues.  Our adversarial
  writing process ensures that the clues also
  challenge computers.}
   \label{fig:ex}
\end{figure}

\subsection{Known Exploits of \qb{} Questions}
\label{subsec:exploits}

Like most \abr{qa} datasets, \qb{} questions are
written for \emph{humans}. Unfortunately, the heuristics that
question authors use to select clues do not always apply to
computers. For example, humans are unlikely to memorize every song in
every opera by a particular composer. This, however, is trivial for a
computer. In particular, a simple \abr{qa} system easily
solves the example in Figure~\ref{fig:ex} from seeing the reference to
``Un Bel Di''. Other questions contain uniquely identifying ``trigger
words''~\cite{harris2006prisoner}. For example, ``martensite'' only appears in questions on
\underline{steel}. For these examples, a \abr{qa} system needs
to understand no additional information other than an if--then rule.

One might wonder if this means that factoid \abr{qa} is thus an
uninteresting, nearly solved research problem.  However, some \qb{}
questions are fiendishly difficult for computers. Many questions have
intricate coreference patterns~\cite{guha15coref}, require reasoning
across multiple types of knowledge, or involve complex wordplay. If we can
isolate and generate questions with these difficult phenemona,
``simplistic'' factoid \abr{qa} quickly becomes non-trivial.

\subsection{Models and Datasets}
\label{subsec:models}

We conduct two rounds of adversarial writing. In the first, authors
attack a traditional information retrieval (\abr{ir}) system.
The \abr{ir} model is the baseline from
a \abr{nips} 2017 shared task on \qb{}~\cite{boydgraber2018nips} based on ElasticSearch~\cite{gormley2015elasticsearch}.

In the second round, authors attack either the \abr{ir} model or
a neural \abr{qa} model. The neural model is 
a bidirectional \abr{rnn} using
the gated recurrent unit architecture~\cite{cho2014gru}.
The model treats \qb{} as classification and predicts the answer entity from a sequence of words
represented as 300-dimensional GloVe
embeddings~\cite{pennington2014glove}. Both models
in this round are trained using an expanded dataset of 
approximately 110,000 \qb{} questions. We expand the second round 
dataset to incorporate more diverse answers (\nansweroptions{} entities
versus 11,000 in round one).

\subsection{Interpreting \qb{} Models}

To help write adversarial questions, we expose
what the model is thinking to the authors.
We interpret models using saliency heat maps: each word of the
question is highlighted based on its importance to the model's
prediction~\cite{ribeiro2016should}.

For the neural model, word importance is the decrease in
prediction probability when a word is
removed~\cite{li2016understanding,wallace2018Neighbors}.
We focus on gradient-based
approximations~\cite{simonyan2013saliency,montavon2017methods}
for their computational efficiency.

\setlength{\abovedisplayskip}{10pt}
\setlength{\belowdisplayskip}{10pt}
To interpret a model prediction on an input sequence of $n$
words~$\mb{w}=\langle\bm{w}_1, \bm{w}_2, \ldots
\bm{w}_n\rangle$, we approximate the classifier $f$ with a linear
function of $w_i$ derived from the first-order Taylor expansion. The
importance of $w_i$, with embedding $\bm{v}_i$, is the derivative
of $f$ with respect to the one-hot vector: 
\begin{equation} \frac{\partial f}{\partial w_i} \
   = \frac{\partial f}{\partial \bm{v}_i}\frac{\partial \bm{v}_i}{\partial w_i} \ 
   = \frac{\partial f}{\partial \bm{v}_i} \cdot \bm{v}_i. 
\end{equation} 
This simulates how model predictions change when a particular word's embedding is set to the zero vector---it approximates word removal~\cite{ebrahimi2017hotflip,wallace2018Neighbors}.

For the \abr{ir} model, we use the ElasticSearch Highlight
\abr{api}~\cite{gormley2015elasticsearch}, which provides word
importance scores based on query matches from the inverted index.

\subsection{Adversarial Writing Interface}

\begin{figure*}[t]
\centering
\includegraphics[width=\textwidth, trim={0cm 7cm 16.5cm 5cm},clip]{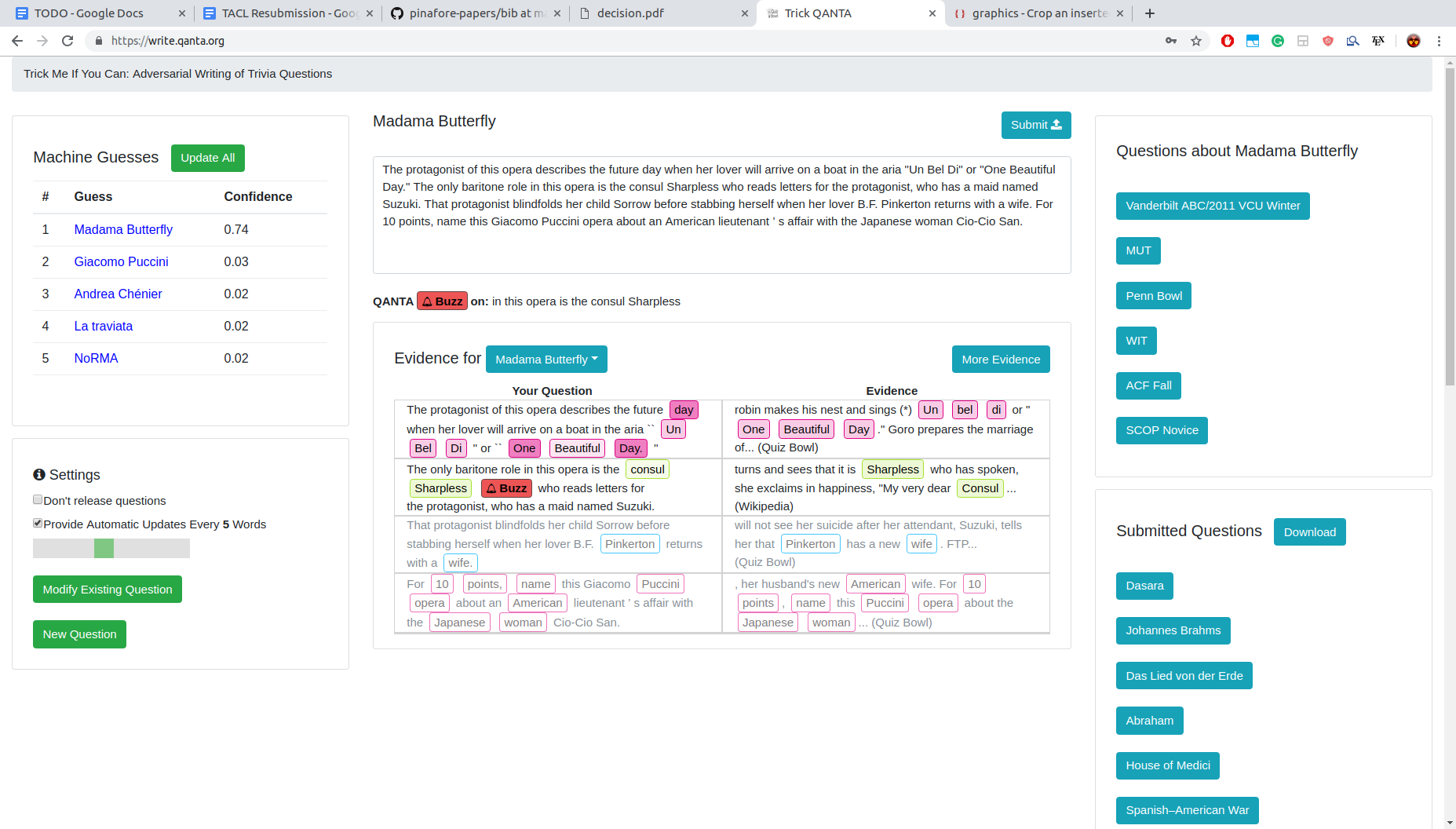}
\caption{The author writes a question (top right), the \abr{qa} system provides
  guesses (left), and explains why it makes those guesses (bottom
  right). The author can then adapt their question to ``trick'' the
  model.}
\label{interface}
\end{figure*}

The authors interact with either the \abr{ir} or \abr{rnn} model through
a user interface\footnote{\url{https://github.com/Eric-Wallace/trickme-interface/}}
(Figure~\ref{interface}). An author writes their question in the upper
right while the model's top five predictions (\textit{Machine
  Guesses}) appear in the upper left. If the top prediction is the
right answer, the interface indicates where in the question
the model is first correct. The goal is to cause the model to
be incorrect or to delay the correct answer position as much as
possible.\footnote{The authors want normal \qb{} questions 
which humans can easily answer by the very end. For popular answers,
  (e.g., \underline{Australia} or \underline{Suez Canal}), writing novel
  final give-away clues is difficult. We thus expect models to often answer correctly
  by the very end of the question.} The words of the
current question are highlighted using the applicable interpretation
method in the lower right (\emph{Evidence}). We do not enforce
time restrictions or require questions to be adversarial: if the
author fails to break the system, they are free to ``give up'' and
submit any question.

The interface continually updates as the author writes.
We track the question edit history to identify recurring model
failures (Section~\ref{sec:limitations}) and understand how
interpretations guide the authors (Section~\ref{sec:help}).

\subsection{Question Authors}

We focus on members of the \qb{} community: they have deep trivia knowledge and craft
questions for \qb{} tournaments~\cite{jennings-06}. We award prizes for questions
read at live human--computer matches (Section~\ref{subsec:live}).

The question authors are familiar with the standard format of \qb{}
questions~\cite{lujan2003writing}. The questions follow a common
paragraph structure, are well edited for grammar, and finish with a
simple ``give-away'' clue. These constraints benefit the adversarial
writing process as it is very clear what constitutes a difficult but
valid question. Thus, our examples go beyond surface level ``breaks''
such as character noise~\cite{belinkov2018synthetic} or syntax
changes~\cite{iyyerscpn2018}. Rather, questions are difficult because
of their semantic content (examples in Section~\ref{sec:limitations}).

\subsection{How an Author Writes a Question}

\AtBeginEnvironment{quote}{\vspace{0.05\baselineskip}}
\AtEndEnvironment{quote}{\vspace{0.05\baselineskip}}
To see how an author might write a question with the interface, we walk
through an example of writing a question's first sentence. The
author first selects the answer to their question from the training
set---\underline{Johannes
Brahms}---and begins:
\begin{quote} Karl Ferdinand Pohl showed this composer some pieces on which
  this composer's Variations on a Theme by Haydn were based. \end{quote}
The \abr{qa} system \emph{buzzes} (i.e., it has enough information to
interrupt and answer correctly) after 
``composer''. The author sees that the name ``Karl
Ferdinand Pohl'' appears in Brahms' Wikipedia page and avoids
 that specific phrase, describing Pohl's
position instead of naming him directly:
\begin{quote} This composer was given a theme called ``Chorale St. Antoni'' by the
  archivist of the Vienna Musikverein, which could have been written
  by Ignaz Pleyel.\end{quote}
This rewrite adds in some additional information (there is a scholarly
disagreement over who wrote the theme and its name), and the \abr{qa} system now incorrectly thinks the answer is
\underline{Fr\'ed\'eric Chopin}.
The user can continue to build on the theme, writing \begin{quote}
   While summering in Tutzing, this composer turned that theme into
   ``Variations on a Theme by Haydn''.  \end{quote}
 Again, the author sees that the system buzzes ``Variations on a
 Theme'' with the correct answer.  However, the author can rewrite it
 in its original German, ``Variationen \"uber ein Thema von Haydn'' to
 fool the system.
 The author continues to
create entire questions the model cannot solve.

\section{A New Adversarially-Authored Dataset}\label{sec:human}

Our adversarial dataset consists of \ntotalquestions{} questions with
\ntotalsentences{} sentences across diverse topics (Table~\ref{table:categories}).\footnote{Data available at \url{http://trickme.qanta.org}.} There are \ntotalIR{} questions  
written against the \abr{ir} system and \ntotalNeural{} against the
neural model by \ntotalwriters{} unique authors.
We plan to hold twice-yearly competitions
to continue data collection.

\begin{table}
\centering \small
\begin{tabular}{lr}
\toprule
Science                        & 17\%                  \\
History                        & 22\%                  \\
Literature                     & 18\%                  \\
Fine Arts                      & 15\%                  \\
Religion, Mythology,           & \multirow{2}{*}{13\%} \\
Philosophy, and Social Science &                       \\
Current Events, Geography,     & \multirow{2}{*}{15\%} \\
and General Knowledge          &                       \\
\midrule
Total Questions & \ntotalquestions{} \\
\bottomrule
\end{tabular}
\caption{The topical diversity of the questions in the \challenge{}
  dataset based on a random sample of 100 questions.}
\label{table:categories}
\end{table}

\subsection{Validating Questions with Quizbowlers}
\label{sec:validation}

We validate that the \challenge{} questions are not of poor quality or too difficult for humans. 
We first automatically filter out questions based on
length, the presence of vulgar statements, or repeated submissions
(including re-submissions from the \qb{} training or evaluation
data). 

We next host a human-only \qb{} event using intermediate
and expert players (former and current collegiate \qb{} players).
We select sixty \challenge{} questions and sixty standard high
school national championship questions, both with the same number of
questions per category (list of categories in Table~\ref{table:categories}).

To answer a \qb{} question, a player interrupts the question: the
earlier the better.  To capture this dynamic, we
record both the average answer
position (as a percentage of the question, lower is better) and 
answer accuracy. We shuffle the regular and \challenge{}
questions, read them to players, and record these two metrics. 

The \challenge{} questions are on average \emph{easier} for humans
than the regular test questions. For the \challenge{} set, humans buzz
with 41.6\% of the question 
remaining and an accuracy of 89.7\%. On the standard questions, humans
buzz with 28.3\% of the question remaining and an accuracy of
84.2\%. The difference in accuracy between the two types of questions
is not significantly different ($p = 0.16$ using Fisher's exact test),
but the buzzing position is earlier for \challenge{}
questions ($p = 0.0047$ for a two-sided $t$-test). We expect the questions
that were not played to be of comparable difficulty because they went through
the same submission process and post-processing. We further explore the
human-perceived difficulty of the \challenge{} questions in Section~\ref{subsec:live}.
\section{Computer Experiments}
\label{sec:experiments}

This section evaluates \abr{qa} systems on the \challenge{} questions. We
test three models: the \abr{ir} and \abr{rnn} models shown in the interface, as well as a Deep Averaging Network~\cite[\abr{dan}]{iyyer2015deep} to evaluate the transferability of the adversarial questions. We break our study into two rounds. The first round consists of \challenge{} questions written against
the \abr{ir} system (Section~\ref{subsec:one}); the second round questions target
both the \abr{ir} and \abr{rnn} (Section~\ref{subsec:two}).

Finally, we also hold live competitions that pit the state-of-the-art Studio Ousia model~\cite{yamada2018studio} against human teams (Section~\ref{subsec:live}).

\begin{figure*}[t!]
\centering  
\includegraphics[width=2\columnwidth]{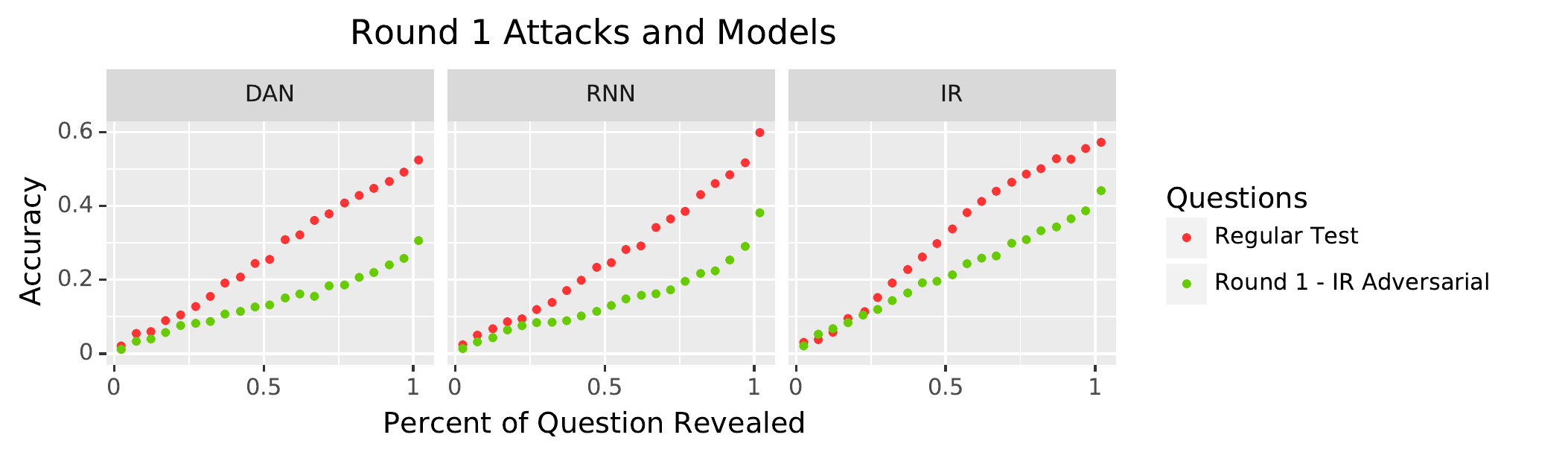}
\caption{The first round of adversarial writing attacks the \abr{ir} model. Like regular test
questions, \challenge{} questions begin with difficult clues that trick the model. However,
the adversarial questions are significantly harder during the crucial middle
third of the question.}
\label{fig:round_one}

\centering
\includegraphics[width=2\columnwidth]{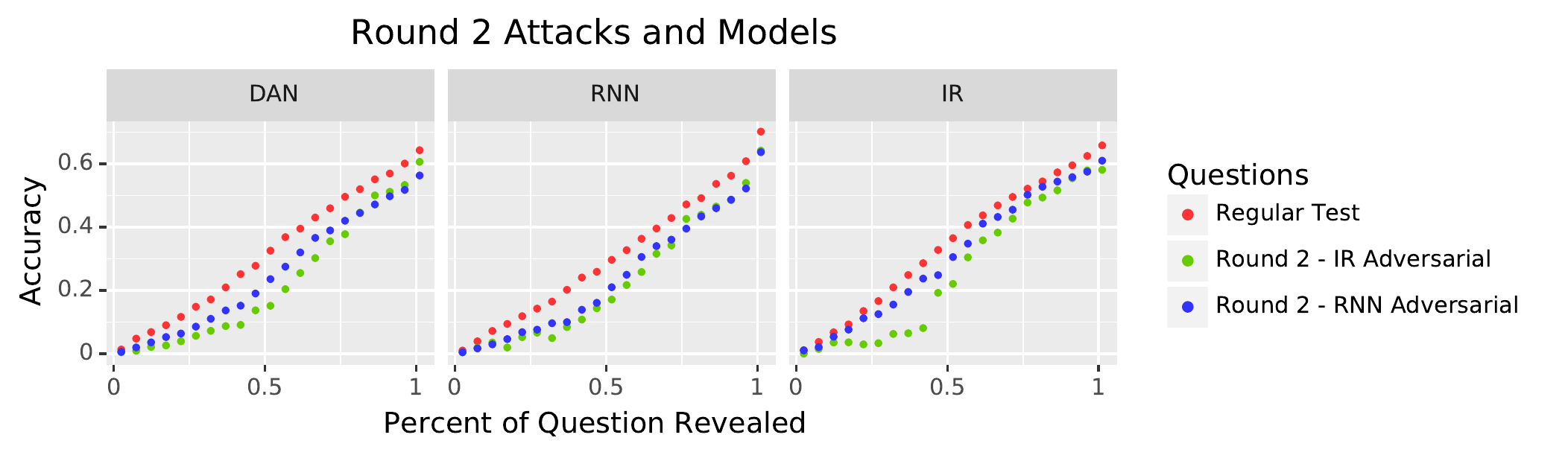}
\caption{The second round of adversarial writing attacks the \abr{ir} and \abr{rnn} models. The questions targeted against the \abr{ir} system degrade the performance of all models. However, the reverse does not hold: the \abr{ir} model is robust to the questions written to fool the \abr{rnn}.}
\label{fig:round_two}
\end{figure*}

\subsection{First Round Attacks: IR Adversarial
Questions Transfer To All Models}\label{subsec:one}

The first round of \challenge{} questions target the \abr{ir} model
and are significantly harder for the \abr{ir},
\abr{rnn}, and \abr{dan} models (Figure~\ref{fig:round_one}). For example,
the \abr{dan}'s accuracy drops from 54.1\% to 32.4\% on the full question
 (60\% of original performance).

For both \challenge{} and original test questions, early
clues are difficult to answer (accuracy about
10\% for the first quarter of the question). However, during the middle third 
of the questions, where buzzes in \qb{} most
frequently occur, accuracy on original test questions rises
 quicker than the \challenge{} ones. For both,
the accuracy rises towards the end as the clues become
``give-aways''.

 \subsection{Second Round Attacks: RNN Adversarial Questions are Brittle}\label{subsec:two}

In the second round, the authors also attack an \abr{rnn} model.
All models tested in the second round are trained on a larger dataset (Section~\ref{subsec:models}).

A similar trend holds for \abr{ir} adversarial questions
in the second round (Figure~\ref{fig:round_two}): a question that tricks the \abr{ir} system also fools
the two neural models (i.e., adversarial examples transfer). For example, the \abr{dan} model
was never targeted but had substantial
accuracy decreases in both rounds.

However, this does not hold for questions written adversarially
against the \abr{rnn} model. On these questions, the neural models
struggle but the 
\abr{ir} model is largely
unaffected (Figure~\ref{fig:round_two}, right).

\subsection{Humans vs. Computer, Live!}
\label{subsec:live}

\begin{figure*}
\centering
\includegraphics[width=2\columnwidth]{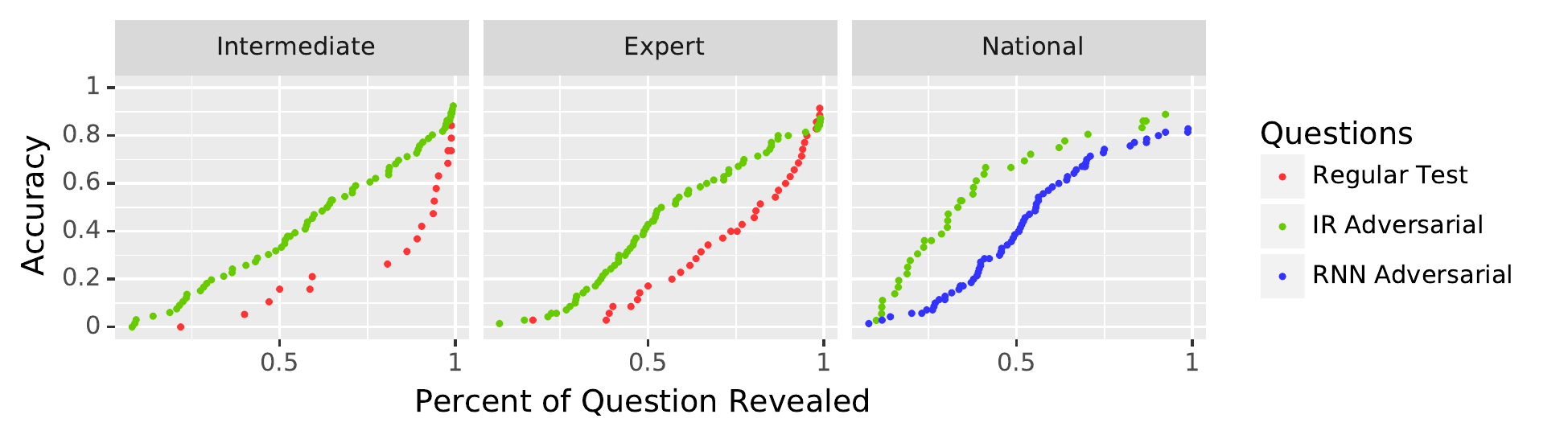}
\caption{Humans find \challenge{} question about as difficult as
  normal questions: rusty weekend
  warriors (\textit{Intermediate}), active players (\textit{Expert}), or
  the best trivia players in the world (\textit{National}).}
\label{fig:human_breakdown}
\end{figure*}

\begin{figure}[t!]
\centering
\includegraphics[width=0.85\columnwidth]{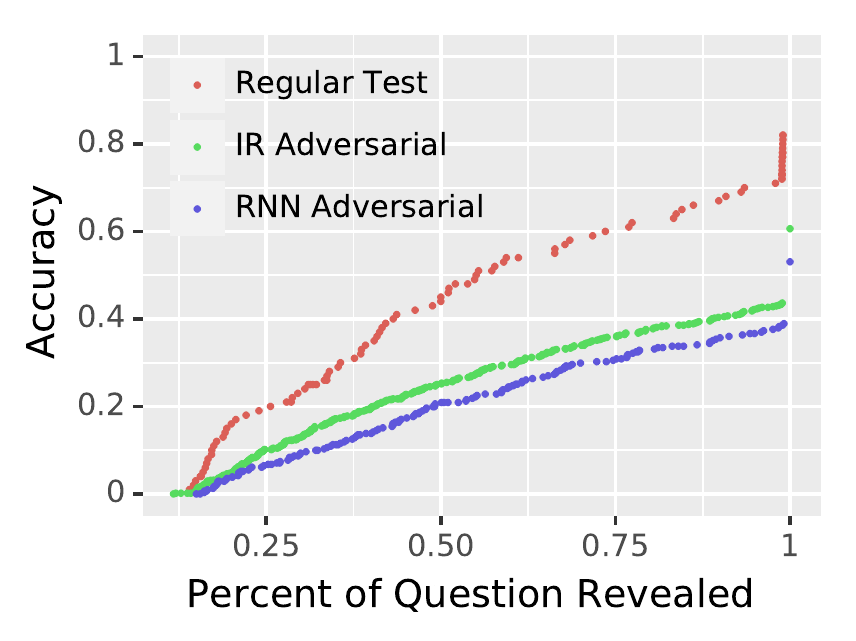}
\caption{The accuracy of the state-of-the-art Studio Ousia model
  degrades on the \challenge{} questions despite never being directly
  targeted. This verifies that our findings generalize beyond the
  \abr{rnn} and \abr{ir} models.}
\label{fig:ikuya_vs_human}
\end{figure}

In the offline setting (i.e., no pressure to ``buzz'' before an opponent) models
demonstrably struggle on the adversarial questions. But, what happens in
standard \qb{}: live, head-to-head 
games? 

We run two live humans vs. computer matches. The first match uses \abr{ir}
adversarial questions in a forty question, tossup-only \qb{} format. 
We pit a human team of national-level \qb{} players against the
Studio Ousia model~\cite{yamada2018studio},
the current state-of-the-art \qb{} system. The model combines neural,
\abr{ir}, and knowledge graph components (details in Appendix~\ref{sec:ousia}), and won the 2017 \abr{nips} shared task, defeating
a team of expert humans 475--200 on regular \qb{} test questions.
Although the
team at our live event was comparable to the \abr{nips}
2017 team, the tables were turned: the human team won handedly 300--30.

Our second live event is significantly larger: seven human teams play
against models on over 400 questions written adversarially against the
\abr{rnn} model. The human teams range in ability from high school
\qb{} players to national-level teams (Jeopardy! champions, Academic
Competition Federation national champions, top scorers in the World
Quizzing Championships). The models are based on either \abr{ir} or
neural methods. Despite a few close games between the weaker human
teams and the models, humanity prevailed in every
match.\footnote{Videos available at \url{http://trickme.qanta.org}.}

Figures~\ref{fig:human_breakdown}--\ref{fig:ikuya_vs_human} summarize
the live match results for the humans and Ousia model, respectively. Humans
and models have considerably different trends in answer accuracy.
Human accuracy on both regular and adversarial questions rises
quickly in the \emph{last half}
of the question (curves in Figure~\ref{fig:human_breakdown}).
In essence, the ``give-away'' clues at the end of questions are easy for humans to answer.

On the other hand, models on regular test questions
do well in the \emph{first half}, i.e.,
the ``difficult'' clues for humans are easier for models (\emph{Regular Test} in Figure~\ref{fig:ikuya_vs_human}).
However, models, like humans, struggle on 
 adversarial questions in the first half.

\section{What Makes Adversarially-authored Questions Hard?}
\label{sec:limitations}

This section analyzes the
\challenge{}
questions to identify the source of their difficulty.

\subsection{Quantitative Differences in Questions}

One possible source of difficulty 
is data scarcity: the answers to adversarial questions rarely
appear in the training set. However, this is not the case;
the mean number of training examples per answer (e.g., \underline{George Washington})
is 14.9 for the adversarial questions versus
16.9 for the regular test data.

Another explanation for question difficulty
is limited 
``overlap'' with the training data, i.e.,
models cannot match $n$-grams
from the training clues. We measure the proportion of test $n$-grams
that also appear in training questions with the same answer
(Table~\ref{table:training_comparison}).
The overlap is roughly equal for unigrams but surprisingly higher for 
adversarial questions' bigrams.
The adversarial questions are also shorter
and have fewer \abr{ne}s.
However, the
proportion of named entities is
roughly equivalent. 

One difference between the questions written against
the \abr{ir} system and the ones written against the \abr{rnn} model
is the drop in \abr{ne}s. The decrease in \abr{ne}s is higher for \abr{ir} adversarial questions, which may explain their generalization: the \abr{rnn}
is more sensitive to changes in phrasing, while the \abr{ir} system is more
sensitive to specific words.

\setlength{\tabcolsep}{4pt}
\begin{table}[t]
\begin{tabular}{p{3.65cm}rr}
 \toprule

& \textbf{Adversarial} & \textbf{Regular}  \\
\midrule
Unigram overlap & 0.40 & 0.37 \\
Bigram overlap & 0.08 & 0.05 \\ 
Longest $n$-gram overlap & 6.73 & 6.87 \\
Average \abr{ne} overlap & 0.38 & 0.46 \\
\hspace{0.5cm} \abr{ir} Adversarial & 0.35 &  \\
\hspace{0.5cm} \abr{rnn} Adversarial & 0.44 &  \\
\midrule
Total Words & 107.1 & 133.5 \\
Total \abr{ne} & 9.1 & 12.5 \\
\bottomrule
\end{tabular}
\caption{The \challenge{} questions have similar $n$-gram overlap to
  the regular test questions. However, the overlap of the
  named entities (\abr{ne}) decreases
   for \abr{ir} Adversarial questions.}
\label{table:training_comparison}
\end{table}

\subsection{Categorizing Adversarial Phenomena}

\begin{table}[h]
\centering \small
\begin{tabular}{lrr}
 \toprule
Composing Seen Clues & 15\%\\
Logic \& Calculations & 5\% \\
Multi-Step Reasoning & 25\% \\
\midrule
Paraphrases & 38\% \\
Entity Type Distractors & 7\%\\
Novel Clues & 26\% \\
\midrule
Total Questions & \ntotalquestions{} \\
\bottomrule
\end{tabular}
\caption{A breakdown of the phenomena in the \challenge{} dataset.}
\label{table:stats}
\end{table}

We next qualitatively analyze \challenge{} questions. We
manually inspect
the author edit logs, classifying questions into six different
phenomena in
two broad categories (Table~\ref{table:stats})
from a random sample of 100 questions, double counting questions into multiple phenomena
when applicable.

\subsubsection{Adversarial Category 1: Reasoning}
\label{sec:compose_knowledge}

\begin{table*}[t]
\centering
\begin{tabular}{p{6.5cm}llp{2cm}}
\hline
         Question    & Prediction & Answer & Phenomenon                  \\ \hline

This man, who died at the Battle of the Thames, experienced a setback when his brother Tenskwatawa's influence over their tribe began to fade. & Battle of Tippecanoe & \underline{Tecumseh} & Composing Seen Clues \\ \hline

This number is one hundred fifty more than the number of Spartans at Thermopylae. & Battle of Thermopylae & \underline{450} & Logic \& Calculations \\ \hline

A building dedicated to this man was the site of the ``I Have A Dream'' speech. & Martin Luther King Jr. & \underline{Abraham Lincoln} & Multi-Step Reasoning \\ \hline

\end{tabular}
\caption{The first category of \challenge{} questions consists of examples that require reasoning. \emph{Answer} displays the correct answer (all models were incorrect). For these examples, connecting the training and \challenge{} clues is simple for humans but difficult for models.}
\label{table:unseen_sample}
\end{table*}

\begin{table*}[t]
\centering
\begin{tabular}{p{1.75cm}p{8.5cm}lp{2cm}}
\hline
         Set & Question    & Prediction      & Phenomenon                  \\ \hline

Training & Name this sociological phenomenon, the \emph{taking of one's own life}. & \underline{Suicide} & \multirow{2}{*}{Paraphrase} \\
Adversarial  & Name this \emph{self-inflicted method of death}. & \underline{Arthur Miller} & \\ \hline

Training &  Clinton played the \emph{saxophone on The Arsenio Hall Show}.  & \underline{Bill Clinton} &  \\ 
Adversarial & He was edited to appear in the film ``Contact''\dots\ For ten points, name this American president who played the \emph{saxophone on an appearance on the Arsenio Hall Show}. & \underline{Don Cheadle} & Entity Type Distractor \\ \hline

\end{tabular}
\caption{The second category of adversarial questions consists of clues that are present in the training data but are written in a distracting manner. \emph{Training} shows relevant snippets from the training data. \emph{Prediction} displays the \abr{rnn} model's answer prediction (always correct on Training, always incorrect on Adversarial).}
\label{table:rewrite_sample}
\end{table*}

The first question category requires reasoning about known clues (Table~\ref{table:unseen_sample}).~\smallskip

\paragraph{Composing Seen Clues:} These questions provide entities with a first-order
relationship to the correct answer. The system must 
triangulate the correct answer by ``filling in the blank''. For
example, the first question of Table~\ref{table:unseen_sample} names the
place of death of Tecumseh. The training data contains a
question about his death reading
``though stiff fighting came from their Native American allies under
Tecumseh, who died at this battle'' (\underline{The Battle of the Thames}). The system must connect these two
clues to answer.~\smallskip

\paragraph{Logic \& Calculations:} These questions require mathematical
or logical operators. For example, the training data
contains a clue about the \underline{Battle of Thermopylae}: ``King
Leonidas and 300 Spartans died at the hands of the Persians''. The
second question in Table~\ref{table:unseen_sample} requires adding
150 to the number of Spartans.~\smallskip 

\paragraph{Multi-Step Reasoning:} This question type requires multiple reasoning
steps between entities. For example, the last question of
Table~\ref{table:unseen_sample} requires a reasoning step
from the ``I Have A Dream'' speech to the Lincoln Memorial and then 	another
reasoning step to reach \underline{Abraham Lincoln}. 

\subsubsection{Adversarial Category 2: Distracting Clues}
\label{sec:changes_language}

The second category consists of circumlocutory clues (Table~\ref{table:rewrite_sample}).~\smallskip

\paragraph{Paraphrases:} A common adversarial modification is to
paraphrase clues to remove exact $n$-gram matches from the training data. This
renders our \abr{ir} system useless but also hurts the neural models. Many of
the adversarial paraphrases go beyond syntax-only changes (e.g., the first row of Table~\ref{table:rewrite_sample}).~\smallskip

\paragraph{Entity Type Distractors:} Whether explicit or implicit in a model, one key component for \abr{qa} is determining the
answer type of the question. Authors take advantage of this
by providing clues that cause the model to select the wrong answer type. For example,
in the second question of Table~\ref{table:rewrite_sample}, the ``lead-in'' clue implies
the answer may be an actor. The \abr{rnn} model answers Don Cheadle in response despite previously
seeing the Bill Clinton ``playing a saxophone'' clue in the training data.~\smallskip

\paragraph{Novel Clues:} Some \challenge{} questions are hard
not because of phrasing or logic but because our models have
not seen these clues.  These questions are easy
to create: users can add \emph{Novel Clues} that---because they are
not uniquely associated with an answer---confuse the models.
While not as linguistically interesting, novel clues are
not captured by Wikipedia or \qb{} data, thus improving the
dataset's diversity.  For example, adding clues about literary
criticism~\cite{hardwick-67,watson-96} to a question about Lillian
Hellman's \underline{The Little Foxes}: ``Ritchie Watson commended
this play's historical accuracy for getting the price for a dozen eggs
right---ten cents---to defend against Elizabeth Hardwick's contention
that it was a sentimental history.'' Novel clues create 
an incentive for models to use
information beyond past questions and Wikipedia.

Novel clues have different effects on \abr{ir} and neural models:
while \abr{ir} models largely ignore them, novel clues can lead
neural models astray.  For example, on a question about \underline{Tiananmen
Square}, the \abr{rnn} model buzzes on
the clue ``World Economic Herald''.  However, adding a novel clue
about ``the history of shaving''
renders the brittle \abr{rnn} unable to buzz on the ``World Economic
Herald'' clue that it was able to recognize before.\footnote{The ``history of shaving'' is a tongue-in-cheek name for a poster
displaying the hirsute leaders of Communist thought. It goes from the bearded Marx and Engels,
to the mustachioed Lenin and Stalin, and finally the clean-shaven Mao.}  This helps
to explain why \challenge{} questions written against the \abr{rnn}
do not stump \abr{ir} models.

\begin{figure}[h]
\centering
\tikz\node[draw=black!40!green,inner sep=1pt,line width=0.3mm,rounded corners=0.1cm]{
\begin{tabular}{p{0.46\textwidth}}
One of these concepts $\ldots$ a \textbf{Hyperbola} is a type of, for ten
points, what shapes made by passing a \textbf{plane} through a namesake solid, \\
\mybox{gitred}{\sout{that also includes the \textbf{ellipse}, \textbf{parabola}?}} \\
\mybox{gitgreen}{whose area is given by one-third Pi r squared} \mybox{gitgreen}{times height?} \\
\emph{Prediction}: \underline{Conic Section} (\Checkmark) $\to$ \underline{Sphere} (\xmark)
\end{tabular}
};
\caption{The interpretation successfully aids an attack
  against the \abr{ir} system. The author removes the phrase containing the words
  ``ellipse'' and ``parabola'', which are highlighted in the interface (shown
  in bold). In its place, they add a phrase which the model associates with the
  answer \underline{Sphere}.}
\label{fig:success} 
\end{figure}

\begin{figure*}
\centering
\includegraphics[width=\textwidth]{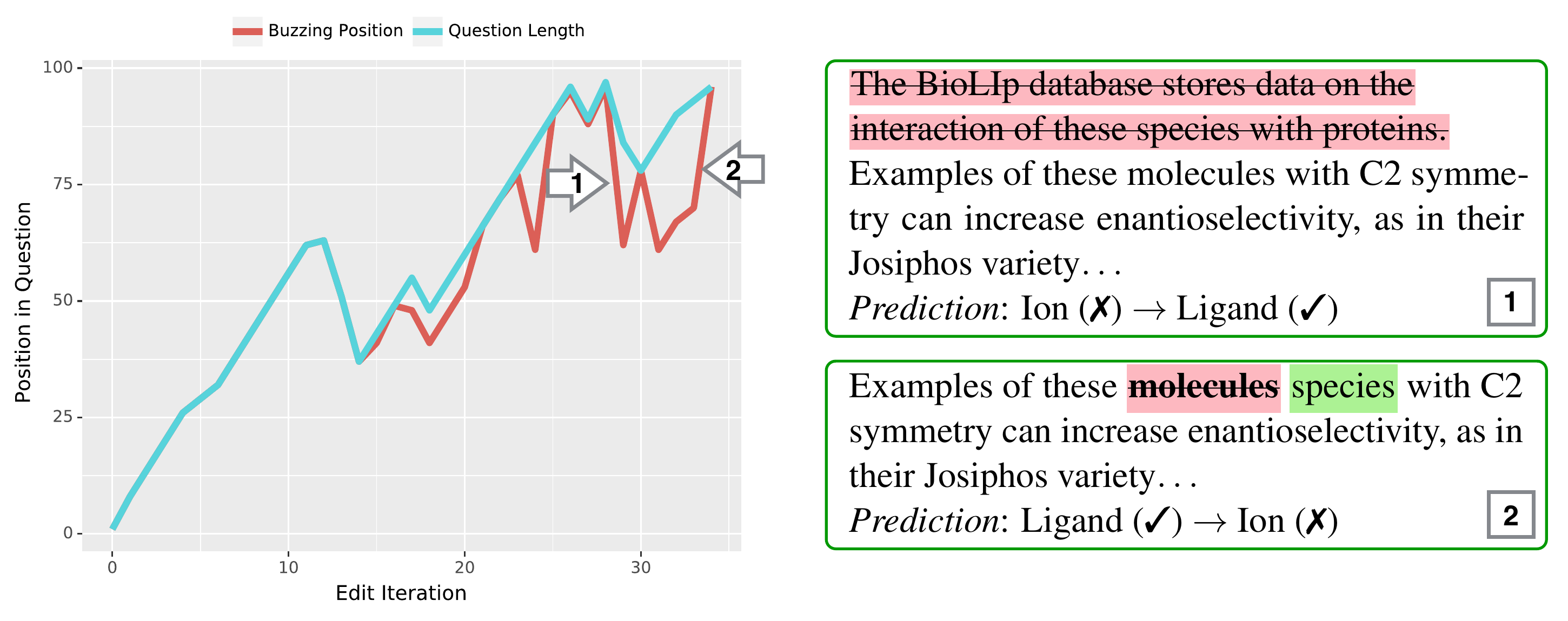}
\caption{The \emph{Question Length} and the
  position where the model is first correct (\textit{Buzzing
    Position}, lower is better) are shown as a question is written. In (\textbf{1}), the author makes a
  mistake by removing a sentence that makes the question easier for the \abr{ir} model. In
  (\textbf{2}), the author uses the interpretation, replacing the
  highlighted word (shown in bold) ``molecules'' with ``species'' to trick
  the \abr{rnn} model.}
\label{fig:edit_example_1}
\end{figure*}

\section{How Do Interpretations Help?}
\label{sec:help}

This section explores how model interpretations help to guide adversarial
authors. We analyze the question edit log, which reflects how authors
modify questions given a model interpretation.

A direct edit of the highlighted words often creates an
adversarial example (e.g., Figure~\ref{fig:success}).
Figure~\ref{fig:edit_example_1} shows a more intricate example. The
left plot shows the \emph{Question Length}, as well as the
position where the model is first correct (\textit{Buzzing Position},
lower is better). We show two adversarial edits. In the first
(\textbf{1}), the author removes the first sentence of the question,
which makes the question \emph{easier} for the model (buzz position
decreases). The author counteracts this in the second edit (\textbf{2}),
where they use the interpretation to craft a targeted modification which
breaks the \abr{ir} model.

However, models are not always this brittle. In
Figure~\ref{fig:failure} (Appendix~\ref{sec:failed}), the interpretation fails to aid an
adversarial attack against the \abr{rnn} model. At each step, the
author uses the highlighted words as a guide to edit targeted portions
of the question yet fails to trick the model. The author gives up and
submits their relatively non-adversarial question.

\subsection{Interviews With Adversarial Authors}

We also interview the adversarial authors who attended our live events. Multiple authors agree that identifying oft-repeated ``stock'' clues was the interface's most useful feature. As one author explained, ``There were clues which I did not think were stock clues but were later revealed to be''. In particular, the author's question about the \underline{Congress of Vienna} used a clue about ``Krak\'ow becoming a free city'', which the model immediately recognized.

Another interviewee was Jordan Brownstein,\footnote{\url{https://www.qbwiki.com/wiki/Jordan_Brownstein}} a national \qb{} champion and one of the best active players, who felt that computer opponents were better at questions that contained direct references to battles or poetry. He also explained how the different writing styles used by each \qb{} author increases the difficulty of questions for computers. The interface's evidence panel allows authors to read existing clues which encourages these unique stylistic choices.

\section{Related Work}
\label{sec:related}

New datasets often allow for a finer-grained analysis of a linguistic
phenomenon, task, or genre.  The \abr{lambada}
dataset~\cite{paperno2016lambada} tests a model's understanding
of the broad contexts present in book passages, while the
Natural Questions corpus~\cite{kwiatkowski2019natural} combs Wikipedia for 
answers to questions that users trust search engines to
answer~\cite{oeldorf2014search}. Other work focuses on natural
language inference, where challenge examples highlight model
failures~\cite{wang2018glue,glockner2018breaking,naik2018stress}.  Our
work is unique in that we use human adversaries to expose model
weaknesses, which provides a diverse set of phenomena (from
paraphrases to multi-hop reasoning) that models cannot solve.

Other work puts an adversary in the data annotation or postprocessing loop.
For instance, \citet{dua2019drop} and \citet{zhang2018record} 
filter out easy questions using a baseline \abr{qa} model,
while \citet{zellers2018swag} use stylistic classifiers to filter
language inference examples. Rather than filtering out easy questions,
we instead use human adversaries to generate hard ones.
Similar to our work, \citet{ettinger2017towards} use human adversaries.
We extend their setting by providing humans with model interpretations to
facilitate adversarial writing. Moreover, we have a
ready-made audience of question writers to generate
adversarial questions.  

The collaborative adversarial writing process reflects
the complementary abilities of humans and computers.
For instance, ``centaur'' chess teams of both a human and a computer are often stronger
than a human or computer alone~\cite{Case2018How}. In Starcraft, 
humans devise high-level ``macro'' strategies, while computers
are superior at executing fast and precise ``micro'' actions~\cite{vinyals2017starcraft}. In \abr{nlp}, computers
aid simultaneous human interpreters~\cite{he2016interpretation} at remembering
forgotten information or translating unfamiliar words.

Finally, recent approaches to adversarial evaluation of
\abr{nlp} models (Section~\ref{sec:recent}) typically
target one phenomenon (e.g., syntactic modifications) and
complement our human-in-the-loop approach.

\section{Conclusion}
\label{sec:future}

One of the challenges of machine learning is knowing
why systems fail.  This work brings together two threads that
attempt to answer this question: visualizations and adversarial
examples.  Visualizations underscore the capabilities of existing models,
while adversarial examples---crafted with the ingenuity of human
experts---show that these models are still
far from matching human prowess. 

Our experiments with both neural and \abr{ir} methodologies show that \abr{qa} models
still struggle with synthesizing clues, handling distracting information,
and adapting to unfamiliar data.  Our \challenge{} dataset is only the first of
many iterations~\cite{ruef16build}. As models improve, future \challenge{}
datasets can elucidate the limitations of next-generation \abr{qa} systems.

While we focus on \abr{qa}, our procedure is applicable to
other \abr{nlp} settings where there is (1) a pool of talented authors
who (2) write text with specific goals. Future research
can look to craft \challenge{} datasets for other \abr{nlp} tasks
that meet these criteria.
\section*{Acknowledgments}

We thank all of the Quiz Bowl players, writers, and judges who helped
make this work possible, especially Ophir Lifshitz and Daniel Jensen.
We also thank the anonymous reviewers and members of the UMD ``Feet
Thinking'' group for helpful comments.
Finally, we would also like to thank Sameer Singh, Matt Gardner,
Pranav Goel, Sudha Rao, Pouya Pezeshkpour, Zhengli Zhao, and Saif
Mohammad for their useful feedback.
This work was supported by \abr{nsf} Grant \abr{iis}-1822494.  Shi
Feng is partially supported by subcontract to Raytheon \abr{bbn}
Technologies by \abr{darpa} award HR0011-15-C-0113, and Pedro
Rodriguez is partially supported by \abr{nsf} Grant \abr{iis}-1409287
(\abr{umd}).
Any opinions, findings, conclusions, or recommendations expressed here
are those of the authors and do not necessarily reflect the view of
the sponsor.

\bibliography{journal-full,eric,jbg}
\bibliographystyle{acl_natbib}

\clearpage
\begin{appendix}
\appendix

\section{Failure of Syntactically Controlled Paraphrase Networks}
\label{sec:scpn}

\begin{table*}[t]
\centering
\begin{tabular}{p{11cm}l}
\hline
         Sentence & Success/Failure Phenomena                  \\ \hline
its types include ``frictional'', ``cyclical'', and ``structural'' & \multirow{2}{*}{Missing Information \xmark} \\
its types include \mybox{gitred}{``frictional'', and structural} & \\ \hline

german author of the sorrows of young werther and a two-part faust & \multirow{2}{*}{Lost Named Entity \xmark} \\
german author of the sorrows of \mybox{gitred}{mr. werther} & \\ \hline

name this elegy on the death of john keats composed by percy shelley & \multirow{2}{*}{Incorrect Clue \xmark} \\
name was this elegy on the \mybox{gitred}{death of percy shelley} & \\ \hline

identify this play about willy loman written by arthur miller & \multirow{2}{*}{Unsuited Syntax Template \xmark}  \\
\mybox{gitred}{so you can identify this work of mr. miller} & \\ \hline

he employed marco polo and his father as ambassadors & \multirow{2}{*}{Verb Synonym \checkmark}  \\ 
he \mybox{gitgreen}{hired} marco polo and his father as ambassadors & \\ \hline

\end{tabular}
\caption{Failure and success cases for \abr{SCPN}. The model fails to create a valid paraphrase of the sentence for  97\% of questions.}
\label{table:scpn}
\end{table*}

We apply the Syntactically Controlled Paraphrase Network~\cite[\abr{SCPN}]{iyyerscpn2018} to \qb{} questions. The model operates on the sentence level and cannot paraphrase paragraphs. We thus feed in each sentence independently, ignoring possible breaks in coreference. The model does not correctly paraphrase most of the complex sentences present in \qb{} questions. The paraphrases were rife with issues: ungrammatical, repetitive, or missing information. 

To simplify the setting, we focus on paraphrasing the shortest sentence from each question (often the final clue). The model still fails in this case. We analyze a random sample of 200 paraphrases: only six maintained all of the original information. 

Table~\ref{table:scpn} shows common failure cases. One recurring issue is an inability to maintain the correct named entities after paraphrasing. In \qb{}, maintaining entity information is vital for ensuring question validity. We were surprised by this failure because \abr{SCPN} incorporates a copy mechanism.

\section{Studio Ousia \qb{} Model}
\label{sec:ousia}

The Studio Ousia system works by aggregating scores from both a neural text classification model and an \abr{IR} system. Additionally, it scores answers based on their match with the correct entity type (e.g., religious leader, government agency, etc.) predicted by a neural entity type classifier. The Studio Ousia system also uses data beyond \qb{} questions and the text of Wikipedia pages, integrating entities from a knowledge graph and customized word vectors~\cite{yamada2018studio}.

\section{Failed Adversarial Attempt}
\label{sec:failed}

Figure~\ref{fig:failure} shows a user's failed attempt to break the neural \qb{} model.

\begin{figure}[h]
\centering
\tikz\node[draw=black!40!red,inner sep=1pt,line width=0.3mm,rounded corners=0.1cm]{
\begin{tabular}{p{0.45\textwidth}}
In his speeches this $\ldots$ As a Senator, \\\mybox{gitred}{\sout{this man supported \textbf{Paraguay} in the}} \\
\mybox{gitred}{\sout{\textbf{Chaco War}, believing \textbf{Bolivia} was backed}}
\mybox{gitred}{\sout{by Standard Oil.}}\\
\mybox{gitgreen}{this man's campaign was endorsed by} \mybox{gitgreen}{Milo Reno and Charles Coughlin.} \\
\emph{Prediction}: \underline{Huey Long} (\Checkmark) $\to$ \underline{Huey Long} (\Checkmark)\\\\
In his speeches this $\ldots$ As a Senator, this man's campaign was endorsed by \textbf{Milo Reno} and \\
\mybox{gitred}{\sout{\textbf{Charles Coughlin}}.} \\
\mybox{gitgreen}{a Catholic priest and radio show host.} \\
\emph{Prediction}: \underline{Huey Long} (\Checkmark) $\to$ \underline{Huey Long} (\Checkmark)\\
\end{tabular}
}; 
\caption{\label{fig:failure} A failed attempt to trick the neural model.  The
author modifies the question multiple times, replacing words suggested by the
interpretation, but is unable to break the system.}
\end{figure}
\end{appendix}

\end{document}